\documentclass[11pt]{article}

\usepackage[preprint]{acl}

\usepackage{times}
\usepackage{latexsym}

\usepackage[T1]{fontenc}

\usepackage[utf8]{inputenc}

\usepackage{microtype}

\usepackage{inconsolata}

\usepackage{graphicx}
\usepackage{tabularx}
\usepackage{booktabs}
\usepackage{subcaption}
\usepackage[most]{tcolorbox}
\tcbset{
  colback=gray!5,
  colframe=gray!60,
  boxrule=0.5pt,
  arc=2pt,
  left=6pt,
  right=6pt,
  top=4pt,
  bottom=4pt,
}
\title{Are LLMs Overkill for Databases?: A Study on the Finiteness of SQL}

\author{
Yue Li\thanks{Corresponding author.}, David Mimno, Unso Eun Seo Jo \\
Cornell University \\
\texttt{\{yl3865, mimno, unsojo\}@cornell.edu}
}

\begin{document}
\maketitle
\begin{abstract} 
Translating natural language to SQL for data retrieval has become more accessible thanks to code generation LLMs. But how hard is it to generate SQL code? While databases can become unbounded in complexity, the complexity of queries is bounded by real life utility and human needs. With a sample of 376 databases, we show that SQL queries, as translations of natural language questions are finite in \textit{practical} complexity. There is no clear monotonic relationship between increases in database table count and increases in complexity of SQL queries. In their template forms, SQL queries follow a Power Law-like distribution of frequency where 70\% of our tested queries can be covered with just 13\% of all template types, indicating that the high majority of SQL queries are predictable. This suggests that while LLMs for code generation can be useful, in the domain of database access, they may be operating in a narrow, highly formulaic space where templates could be safer, cheaper, and auditable.
\end{abstract} 


\section{Introduction}
The text-to-SQL task predates the rise of language models \cite{woods1973progress, harris1977robot}, but is currently dominated by LLM-based approaches \cite{li2025agent}. 
For instance, on the premier BIRD text-to-SQL benchmark, almost all of the top ranking methods on the public site are either agent-based \cite{pourreza2024chase} or LLM-based \cite{longdata2025sql}.
While these approaches are effective, they can get expensive, costing several dollars in token-count per query, and potentially unreliable, whereas enterprise database queries are highly sensitive to efficiency and security.
To what extent is this code boilerplate, predictable, or templatizable?
In this work, we probe whether full code generation is really necessary for SQL applications.



We explore the \textit{practical} boundaries of SQL through empirical experiments. We curate 376 database schemas from benchmarks and open source database repository, \textit{drawSQL}\footnote{ \href{https://drawsql.app/templates}{drawsql.app/templates}} and generate over 20,000 natural language questions (NLQ) and matching SQL queries based on schema designs. Using counts of proxies for complexity such as JOIN clause count, subclause count, and length of query, we show that there is no monotonic relationship between increase in table count of databases and increase in proxy count. We observe there is a ``ceiling'' of proxy-based complexity in SQL queries regardless of the complexity and size of the database schema design. 

\begin{figure}[t]
\centering
\includegraphics[width=\columnwidth]{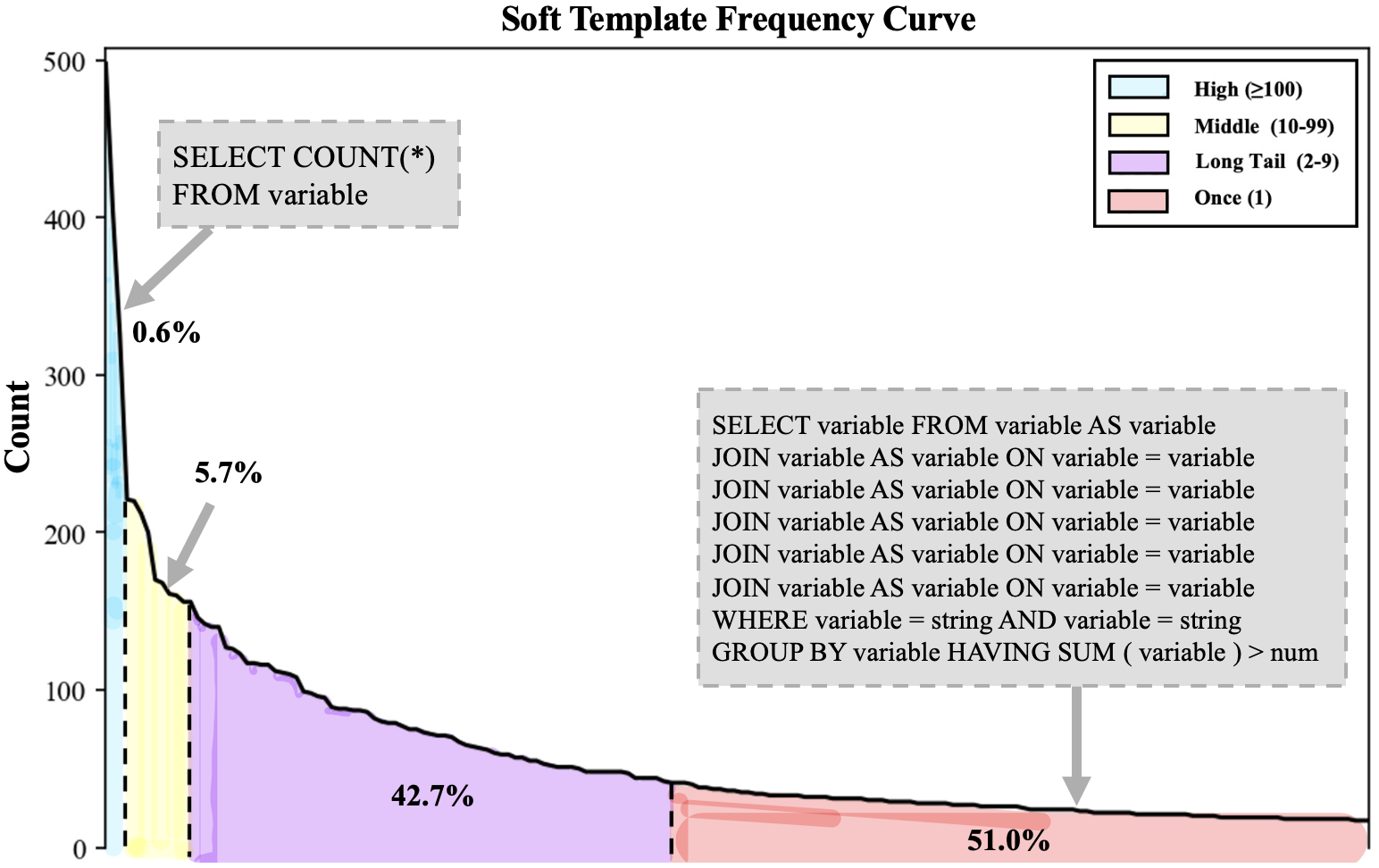}
\caption{Frequency curve of soft templates.}
\label{fig:figure1}
\end{figure}

When templated and analyzed by frequency, SQL queries exhibit a power-law-like distribution (Figure~\ref{fig:figure1}). We find that the top 13\% of templates account for 90\% of all queries: only 7 templates cover 10\%, and about 40 cover 30\%. High-frequency templates tend to be less complex (e.g., fewer JOINs), suggesting that relatively simple structures suffice to cover most user questions. Overall, a few hundred templates are sufficient to capture the vast majority of database queries.

\section{Related Work}
Text-to-SQL has long been studied as a semantic parsing task using deep learning models \cite{zhong2017seq2sql, xu2017sqlnet}. Recent advances in LLMs have shifted this paradigm, with state-of-the-art systems adopting LLM-driven pipelines with iterative reasoning, self-correction, and agent-based workflows \cite{pourreza2023dinsql}. 

While LLM-based SQL generation has significantly improved accuracy, limited work has examined the empirical \textit{science} of database queries. To our knowledge, this is the first work to explore the boundaries of SQL queries, enabled by the recent availability of large-scale, high-quality AI-generated SQL data that allows systematic analysis of query patterns and user intents.
\section{Methodology}
\subsection{Database and Query Collection}



We collect NLQ--SQL pairs and database schemas from online benchmarks and an open source repository, \textit{drawSQL}. We use database schemas and gold label queries from classic text-to-SQL benchmarks such as Spider~1.0 \cite{yu2018spider}. We take schema designs from \textit{drawSQL} and construct NLQ--SQL pairs using Claude-Sonnet-4.6 \cite{anthropic2026sonnet} with varying levels of difficulty. Table~\ref{tab:schema_stats} shows the statistics of database schemas collected from these sources. We manually verify all generated NLQ--SQL pairs to ensure semantic correctness and executable validity. To ensure consistency in template extraction and metric computation, we restrict all SQL queries to SQLite dialect. 


\begin{table}[t]
\centering
\small
\resizebox{\columnwidth}{!}{
\begin{tabular}{lccc}
\toprule 
\textbf{Source} & \textbf{\#DB} & \textbf{\#T/DB} & \textbf{\#Q} \\
\midrule
Bird23-train-filtered~\cite{li2023bird} & 69 & 7.57 & 6,601 \\
Spider~1.0~\cite{yu2018spider} & 196 & 5.15 & 11,245 \\
Spider~2.0-lite~\cite{lei2024spider} & 15 & 14.87 & 287 \\
KaggleDBQA~\cite{lee2021kaggledbqa} & 8 & 2.12 & 244 \\
drawSQL & 88 & 14.34 & 2,112 \\
\textbf{Overall} & \textbf{376} & \textbf{8.07} & \textbf{20,489} \\
\bottomrule
\end{tabular}}
\caption{Statistics of database schemas collected from text-to-SQL benchmarks and drawSQL. \#DB indicates the number of databases from each data source, \#T/DB indicates the average number of tables per database, and \#Q indicates the number of SQL queries per source. For drawSQL, we manually generate three difficulty levels of NLQs: easy, medium, and difficult. Appendix~\ref{sec:NLQs-level} provides concrete examples.}
\label{tab:schema_stats}
\end{table}

\subsection{SQL Templates Extraction}
To analyze the frequency of repeat SQL queries by ``type'', we templatize or generalize the queries with uniform methods. We have two different categories of SQL templates: (1) \textbf{hard templates} keep more distinctive traits, and (2) \textbf{soft templates} uniformize more strictly to allow for looser categorization.

\paragraph{Type 1: Hard Templates.} We construct hard SQL templates by abstracting dataset-specific
identifiers and literals from SQL queries while preserving their structural and logical forms. This process normalizes queries into canonical templates that represent the underlying SQL pattern. Hard templates have stricter and more detailed extraction rules to transform SQL queries into SQL templates. We keep entity types such as table, column, and alias names as placeholders. Appendix~\ref{sec:hard templates-appendix} shows detailed hard templates extraction rules.

\paragraph{Type 2: Soft Templates.} In addition to the masking presented by hard templates, we further generalize the variables and focus on preserving the underlying SQL structure and keywords. Function-specific keywords such as alias and table are all uniformly treated as variables. With this transformation, some distinct ``hard'' templates could be mapped to the same soft template. Appendix~\ref{sec:soft templates-appendix} shows detailed soft template extraction rules.

\paragraph{Example: Template Generation Procedure.}
The original SQL query is translated into templates in stages: \textit{SQL $\rightarrow$ Hard Template $\rightarrow$ Soft Template}.

\begin{quote}
\textbf{SQL Query}
\begin{verbatim}
SELECT c.name 
FROM customers AS c
JOIN orders AS o ON c.id = o.customer_id
WHERE o.amount > 100 
ORDER BY o.amount DESC LIMIT 10
\end{verbatim}

\textbf{Hard Template}
\begin{verbatim}
SELECT table_alias0.col_name 
FROM table_name AS table_alias0
JOIN table_name AS table_alias1
ON table_alias0.col_name = t
able_alias1.col_name
WHERE table_alias1.col_name > num
ORDER BY table_alias1.col_name 
DESC LIMIT num
\end{verbatim}

\textbf{Soft Template}
\begin{verbatim}
SELECT variable
FROM variable AS variable
JOIN variable AS variable
ON variable = variable
WHERE variable > num
ORDER BY variable DESC LIMIT num
\end{verbatim}
\end{quote}

This example illustrates how hard templates preserve precise alias and schema structure, while soft templates collapse identifier roles into a single variable token. Appendix~\ref{sec:templates-appendix} presents more hard and soft template extraction examples.



\subsection{Proxies for Complexity}
One way to quantify ``complexity'' of SQL queries is to count the appearance of certain proxy traits. We identify 6 proxies used to characterize SQL query complexity. These include structural properties of queries (number of tables, joins, subqueries, and maximum nesting depth) as well as analytical operations such as aggregations with \texttt{GROUP BY} and advanced SQL features (e.g., window functions, percentile functions, \texttt{FILTER}, set operations, and CTEs). Table~\ref{sql-proxy-table} provides the full list of all proxies.

\begin{table*} 
  \centering
  \small
  \begin{tabular}{l p{10cm}}
    \hline
    \textbf{Proxy} & \textbf{Definition} \\
    \hline
    Num\_tables & \# of distinct tables referenced (i.e., number of data sources involved) \\
    
    Num\_joins & \# of JOIN operations (capturing cross-table relational reasoning) \\
    
    Num\_subqueries & \# of nested subqueries (capturing hierarchical reasoning) \\
    
    Max\_nesting\_depth & Maximum depth of nested queries (i.e., levels of embedding) \\
    
    Num\_aggs\_plus\_group\_by & \# of aggregation operations (e.g., COUNT, SUM) and GROUP BY clauses \\
    
    Advanced\_feature\_count & \# of advanced constructs (e.g., window functions, FILTER, set operations, CTEs) \\
    
    \hline
  \end{tabular}
  \caption{\label{sql-proxy-table}
    Structural proxies for SQL query complexity, capturing aspects such as multi-table interactions, hierarchical structure, and advanced analytical operations.
  }
\end{table*}

\begin{table}[t]
\centering
\caption{Summary statistics and peak characteristics of six SQL complexity proxies.}
\label{tab:metrics}

\resizebox{\columnwidth}{!}{
\begin{tabular}{lccccc}
\toprule
\textbf{Proxy} & \textbf{Median} & \textbf{Average} & \textbf{Min} & \textbf{Max} & \textbf{Peak Value} \\
\midrule
Num\_tables & 2.00 & 1.73 & 1 & 27 & 1.94 \\
Num\_joins & 1.00 & 0.73 & 0 & 26 & 0.90 \\
Num\_subqueries & 0.00 & 0.11 & 0 & 7 & 0.22 \\
Max\_nesting\_depth & 1.00 & 0.62 & 0 & 5 & 0.74 \\
Num\_aggs\_plus\_group\_by & 0.00 & 0.74 & 0 & 9 & 0.81 \\
Advanced\_feature\_count & 0.00 & 0.06 & 0 & 6 & 0.17 \\
\bottomrule
\end{tabular}
}
\end{table}

\section{Results}
\begin{figure}[t]
\centering
\includegraphics[width=\columnwidth]{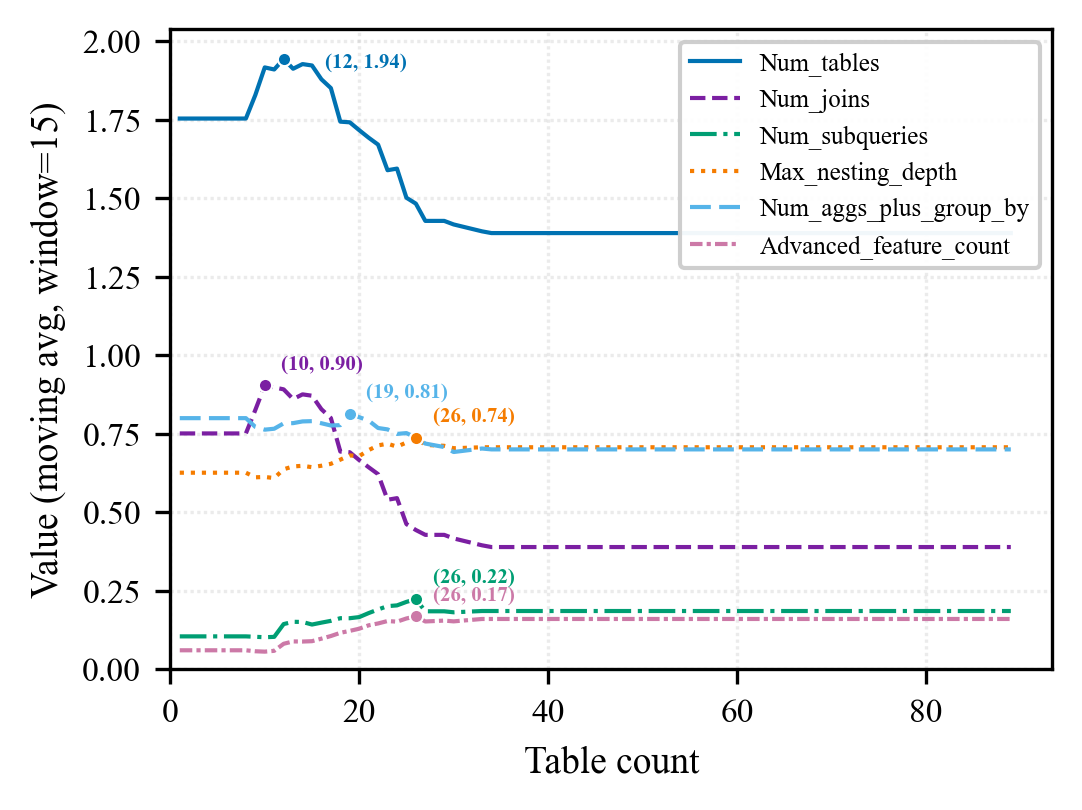}
\caption{Moving average values of six proxies (window size = 15).}
\label{fig:win15}
\end{figure}

\paragraph{Non-monotonic SQL Query Complexity Trend}
There is no observable monotonic relationship between increase in database schema complexity and increase in SQL query complexity. SQL queries do not grow more and more complex as database schemas do. 
Figure~\ref{fig:win15} presents the moving average values of six proxies (window size = 15). All curves exhibit a similar pattern: they increase to some \textit{peak value} at a \textit{breaking point}, and then gradually stall or decline, suggesting that the SQL query complexity measured by these proxies does not continuously grow with the number of tables in the database. Instead, the complexity of SQL queries appears to follow a bounded trend rather than monotonically increasing as database size grows. Table~\ref{tab:spearman_proxy} in Appendix~\ref{appen-Spearman} reports the Spearman Correlation results, showing the absence of a monotonic relationship between the proxies and the table count. 

Table~\ref{tab:metrics} summarizes the statistics of the six proxies across all SQL queries. Most of the proxies appear on average less than once per query. The count of table variables is higher: there are queries that include as many as 27 table variables. But the average table count seems bounded at around 2.

\paragraph{Power Law-like Distribution of Template Frequency.}

When templatized and grouped by frequency, the 20,489 queries follow a near Power Law distribution. The hard and soft templates both exhibit long-tail distributions: a small number of templates appear very frequently while most templates occur only a few times or once. Figure~\ref{fig:frequency_curve} in Appendix~\ref{sec:freq} shows the long-tail pattern. This indicates a limited set of templates dominate while the majority of templates have more specialized SQL structures. The Power Law hypothesis is rejected by bootstrap goodness-of-fit test~\cite{clauset2009power} ($p \approx 0$), but the curves are Power Law-like in the tail.




\begin{figure}[t]
\centering

\begin{subfigure}{\linewidth}
\centering
\includegraphics[width=\linewidth]{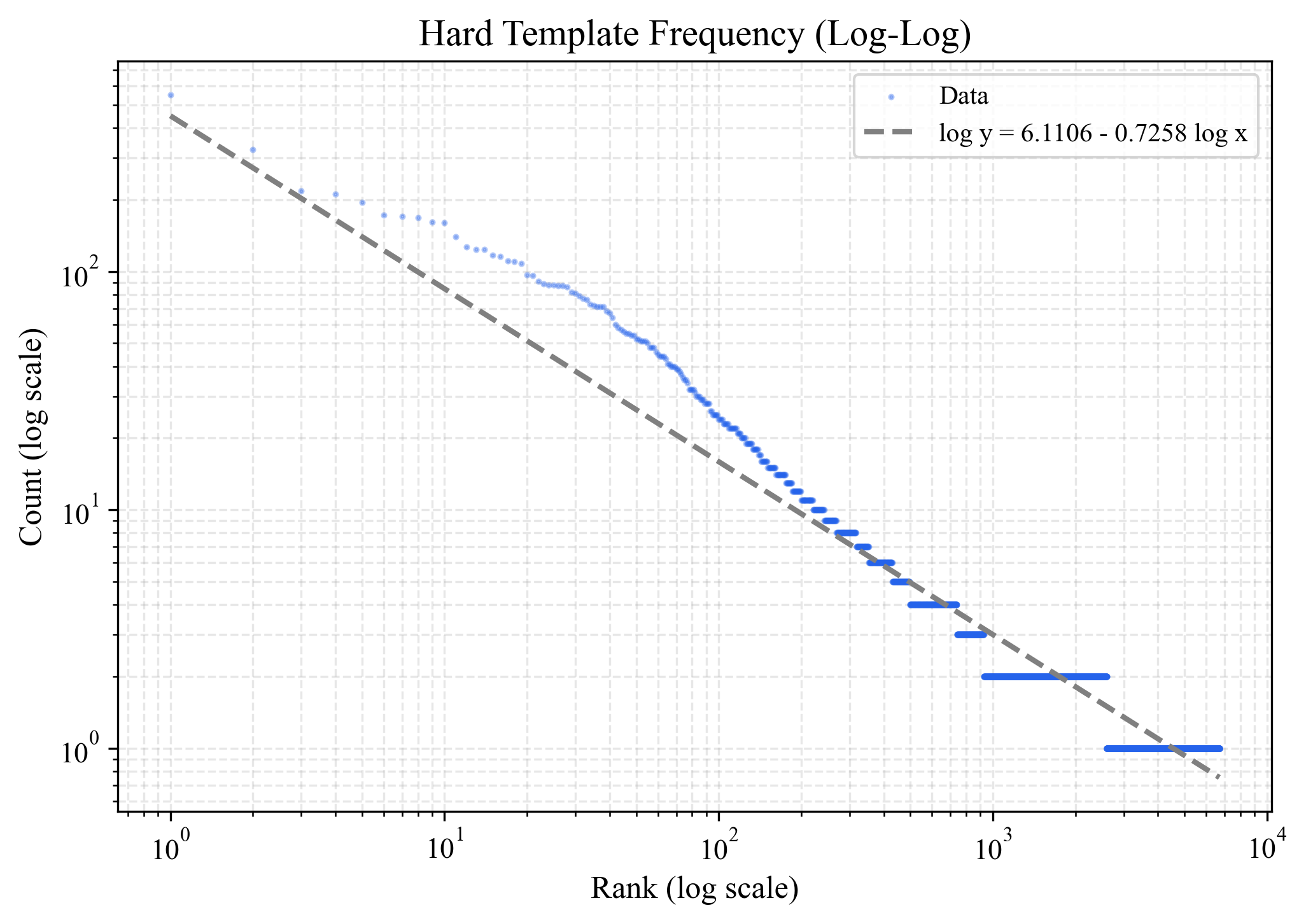}
\caption{Log--log plot of Hard Templates}
\end{subfigure}
\begin{subfigure}{\linewidth}
\centering
\includegraphics[width=\linewidth]{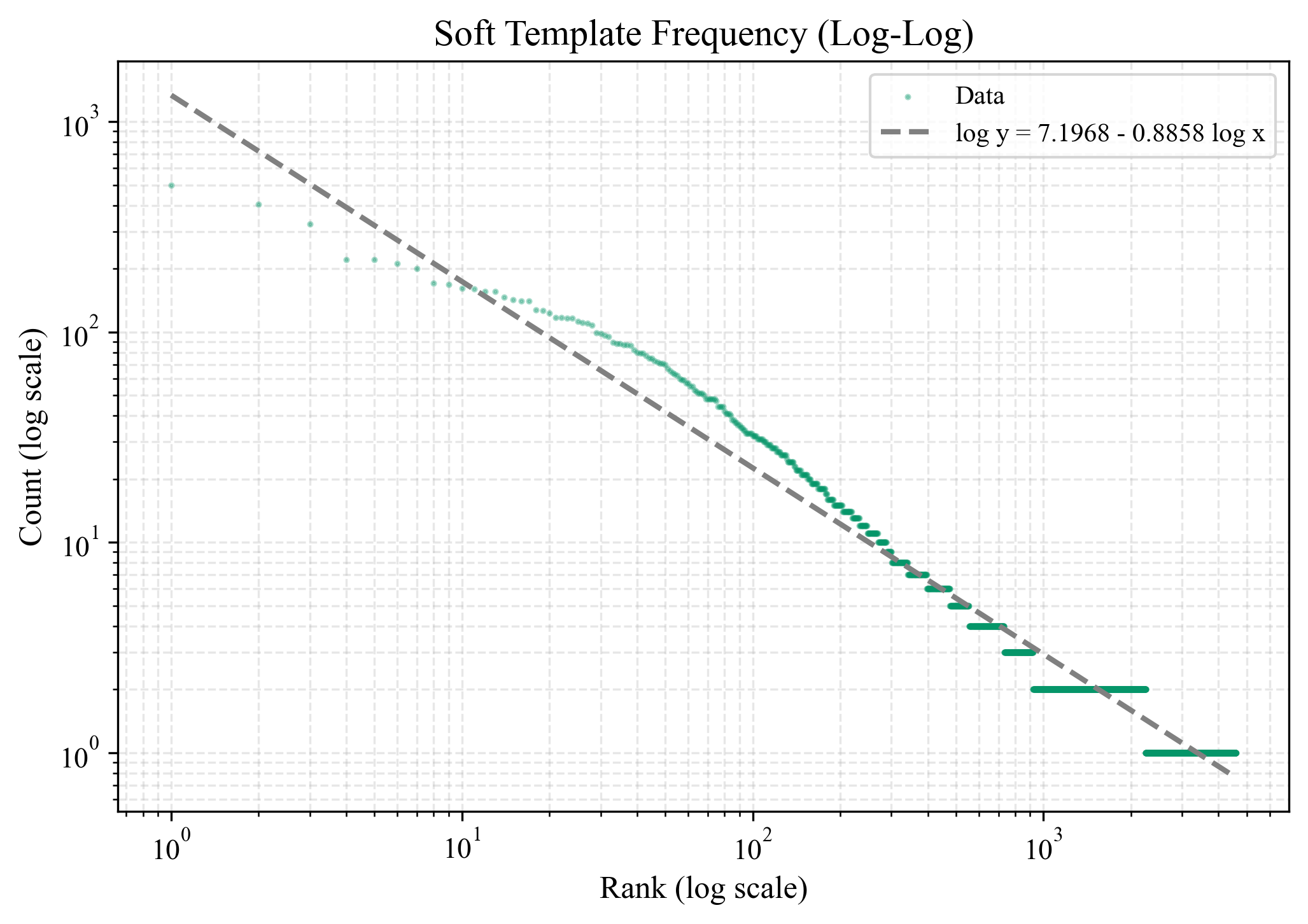}
\caption{Log--log plot of Soft Templates}
\end{subfigure}
\caption{Log--log plots for hard and soft templates.}
\label{fig:log_plots}
\end{figure}

\begin{table}[t]
\centering
\caption{Number of templates required to cover different proportions of SQL queries. 
Numbers in parentheses denote the cumulative percentage of templates.}
\label{tab:coverage}
\resizebox{\columnwidth}{!}{
\begin{tabular}{lcccccc}
\toprule
\textbf{Template Type} & \multicolumn{6}{c}{\textbf{Query Coverage (\%)}} \\
\cmidrule(lr){2-7}
 & 10 & 30 & 50 & \textbf{70} & 90 & 100 \\
\midrule
Hard & 9 (0.13\%) & 58 (0.87\%) & 306 (4.57\%) & \textbf{1616 (24.12\%)} & 4677 (69.82\%) & 6699 (100\%)\\
Soft & 7 (0.15\%) & 42 (0.92\%) & 140 (3.05\%) & \textbf{605 (13.19\%)} & 2565 (55.92\%) & 4587 (100\%)\\
\bottomrule
\end{tabular}
}
\end{table}

We plot hard and soft template frequencies on log-log scale in figure~\ref{fig:log_plots} showing an approximately linear trend. 
We draw fitted lines following the linear representation of the Power Law distribution: $\log P(x) = -\alpha \log x + C$ ,where $C$ is a constant. The fitted lines for hard and soft template frequency log--log scale distributions are $\log y = - 0.7258 \log x + 6.1106$ (hard) and $\log y = - 0.8858 \log x + 7.1968$ (soft), respectively. Not surprisingly, soft templates follow a steeper distribution, where the ``top'' frequency templates have higher counts. Table~\ref{tab:frequency} in Appendix~\ref{tab:frequency} shows soft templates have 28 templates with more than 100 counts whereas hard templates have just 19. Both distributions have long tails of single appearance queries making up 60\% of the hard template distribution.

\paragraph{70\% SQL queries can be covered using only 13.19\% soft templates.} Table~\ref{tab:coverage} shows that a small number of templates can cover a disproportionate volume of queries with the appropriate schema linking of table and column names. The Power Law effect is more dramatic with soft templates, where up to 70\% of all queries are derived from about 13\% of templates. It takes only about 600 soft templates to cover 70\% of 20,489 queries. The most frequent 140 templates cover about 50\% of all queries. The two most frequent soft templates are \texttt{SELECT variable FROM variable WHERE variable = string} and \texttt{SELECT COUNT(*) FROM variable}. For example, the former corresponds to NLQ such as ``List all users whose status is `active`,'' while the latter corresponds to NLQ such as ``How many users are there in total?''

Table~\ref{tab:metric_by_freq} in Appendix~\ref{appen: avg_metric} also shows the average proxy count of complexity negatively correlates with frequency as we expected. The simpler templates are more likely to be used in more contexts.

\section{Conclusion}
Relational and other databases are the middle ground of data storage and organization between human understanding and computational efficiency. SQL was proposed in the 1970s as a way to bridge the gap between human natural language and computer execution. LLMs have emerged as the additional layer over this with its seemingly magical code generation capabilities. But our experiments show LLMs code generation in the SQL domain may not be as impressive as they operate under a ceiling of complexity where just a few hundred templates cover the vast majority of query cases. 

\section*{Limitations}

One limitation of template-based analysis is that SQL queries do not have canonical forms and many different queries can be isomorphic. In other words, it is possible to generate code that looks completely different, and therefore have different templates, but retrieve the same ``answer'' or table result. We experimented with ways of homogenizing the overall style of generated queries to test different levels of maximizing template frequency, but ultimately decided that we wanted to test the queries (and resulting templates) in the most unadulterated, as-found forms. 

\section*{Acknowledgments}
We thank the anonymous reviewers and area chairs for their valuable feedback.

\bibliography{custom}

\appendix

\section{Examples of Generated NLQs by Difficulty Level}
\label{sec:NLQs-level}
We use the \textit{cachet} database from the drawSQL dataset as an illustrative example, and present NLQ--SQL pairs across three difficulty levels: easy, medium, and difficult.

\subsection{Example 1 (Easy, cachet).}
\textbf{NLQ:} \\
How many subscribers are there in total? \\
\textbf{SQL Query:}
\begin{verbatim}
SELECT COUNT(*) 
FROM subscribers;
\end{verbatim}

\subsection{Example 2 (Medium, cachet).}
\textbf{NLQ:} \\
How many activity log entries do not have a matching component? \\
\textbf{SQL Query:}
\begin{verbatim}
SELECT COUNT(*) 
FROM actions a
LEFT JOIN components b 
ON a.taggable_id = b.id
WHERE b.id IS NULL;
\end{verbatim}

\subsection{Example 3 (Hard, cachet).}
\textbf{NLQ:} \\
For each class, count the number of activity log entries, and return only those classes whose count is above the average count across all classes. Sort the results in descending order of the count. \\
\textbf{SQL Query:}
\begin{verbatim}
WITH grouped AS (
    SELECT class_name, COUNT(*) AS cnt
    FROM actions
    GROUP BY class_name
)
SELECT *
FROM grouped
WHERE cnt > (
    SELECT AVG(cnt) FROM grouped
)
ORDER BY cnt DESC;
\end{verbatim}

\section{Templates Extraction Examples}
\label{sec:templates-appendix}

To illustrate the template abstraction process, we present several
examples showing the transformation pipeline from SQL queries to
hard templates and soft templates.

\paragraph{Example 1: Simple Selection}

\begin{quote}
\textbf{SQL}
\begin{verbatim}
SELECT name FROM employees
WHERE salary > 50000
\end{verbatim}

\textbf{Hard Template}
\begin{verbatim}
SELECT col_name FROM table_name
WHERE col_name > num
\end{verbatim}

\textbf{Soft Template}
\begin{verbatim}
SELECT variable FROM variable
WHERE variable > num
\end{verbatim}
\end{quote}

\paragraph{Example 2: Join with Aliases}

\begin{quote}
\textbf{SQL}
\begin{verbatim}
SELECT T1.name
FROM employees AS T1
JOIN departments AS T2
ON T1.dept_id = T2.id
WHERE T2.location = 'NY'
\end{verbatim}

\textbf{Hard Template}
\begin{verbatim}
SELECT table_alias0.col_name
FROM table_name AS table_alias0
JOIN table_name AS table_alias1
ON table_alias0.col_name = 
table_alias1.col_name
WHERE table_alias1.col_name = string
\end{verbatim}

\textbf{Soft Template}
\begin{verbatim}
SELECT variable
FROM variable AS variable
JOIN variable AS variable
ON variable = variable
WHERE variable = string
\end{verbatim}
\end{quote}

\paragraph{Example 3: Aggregation}

\begin{quote}
\textbf{SQL}
\begin{verbatim}
SELECT department, COUNT(*)
FROM employees
GROUP BY department
ORDER BY COUNT(*) DESC
LIMIT 5
\end{verbatim}

\textbf{Hard Template}
\begin{verbatim}
SELECT col_name, COUNT(*)
FROM table_name
GROUP BY col_name
ORDER BY COUNT(*) DESC
LIMIT num
\end{verbatim}

\textbf{Soft Template}
\begin{verbatim}
SELECT variable, COUNT(*)
FROM variable
GROUP BY variable
ORDER BY COUNT(*) DESC
LIMIT num
\end{verbatim}
\end{quote}

\paragraph{Example 4: Subquery}

\begin{quote}
\textbf{SQL}
\begin{verbatim}
SELECT name
FROM employees
WHERE salary >
    (SELECT AVG(salary) FROM employees)
\end{verbatim}

\textbf{Hard Template}
\begin{verbatim}
SELECT col_name
FROM table_name
WHERE col_name >
    (SELECT AVG(col_name) 
    FROM table_name)
\end{verbatim}

\textbf{Soft Template}
\begin{verbatim}
SELECT variable
FROM variable
WHERE variable >
    (SELECT AVG(variable) FROM variable)
\end{verbatim}
\end{quote}

\paragraph{Example 5: CTE with Join and Aggregation}

\begin{quote}
\textbf{SQL}
\begin{verbatim}
WITH dept_avg AS (
  SELECT dept_id, AVG(salary) AS avg_salary
  FROM employees
  GROUP BY dept_id
)
SELECT d.name
FROM departments d
JOIN dept_avg a
ON d.id = a.dept_id
WHERE a.avg_salary > 70000
\end{verbatim}

\textbf{Hard Template}
\begin{verbatim}
WITH CTE0 AS (
  SELECT col_name, AVG(col_name) 
  AS column_alias0
  FROM table_name
  GROUP BY col_name
)
SELECT table_alias0.col_name
FROM table_name AS table_alias0
JOIN CTE0 AS table_alias1
ON table_alias0.col_name = 
table_alias1.col_name
WHERE table_alias1.column_alias0 > num
\end{verbatim}

\textbf{Soft Template}
\begin{verbatim}
WITH CTE0 AS (
  SELECT variable, AVG(variable) 
  AS variable
  FROM variable
  GROUP BY variable
)
SELECT variable
FROM variable AS table_alias0
JOIN CTE0 AS table_alias1
ON variable = variable
WHERE variable > num
\end{verbatim}
\end{quote}

\section{Complete Hard Templates Extraction Rules}
\label{sec:hard templates-appendix}
The generation procedure consists of the following steps:

\begin{enumerate}
\item \textbf{Preprocessing.}
We remove comments (e.g., \texttt{--}, \texttt{/* */}) and blank lines
from the SQL query and perform case-insensitive matching to standardize
the input representation.

\item \textbf{Literal Abstraction.}
All literal values are replaced with typed placeholders.
Numeric constants are mapped to \texttt{num}, string literals to
\texttt{string}, date literals to \texttt{date}, boolean values to
\texttt{boolean}, and \texttt{NULL} values to \texttt{others}.
This step removes value-level variability while preserving type signals.

\item \textbf{Schema-aware Identifier Replacement.}
Using the database schema, table names are replaced with
\texttt{table\_name} and column references with \texttt{col\_name}.
Identifiers that do not appear in the schema are mapped to
special placeholders such as \texttt{new\_table}, \texttt{new\_view},
or \texttt{new\_column} depending on their SQL context.

\item \textbf{Alias Normalization.}
Table aliases introduced in \texttt{FROM} or \texttt{JOIN} clauses are
normalized as \texttt{table\_alias0}, \texttt{table\_alias1}, etc.,
according to their order of appearance.
Column aliases defined via \texttt{AS} are similarly replaced with
\texttt{column\_alias0}, \texttt{column\_alias1}, etc.

\item \textbf{Qualified Reference Resolution.}
For qualified expressions (e.g., \texttt{alias.column}), the qualifier is
replaced by its normalized form (e.g., \texttt{table\_name} or
\texttt{table\_aliasN}), while the column component is mapped to
\texttt{col\_name} or \texttt{new\_column}.
\end{enumerate}

The resulting SQL statement forms a hard template that preserves the structural semantics of the original query while abstracting away dataset-specific details.

\section{Complete Soft Templates Extraction Rules}
The generation procedure consists of the following steps:
\label{sec:soft templates-appendix}
\begin{enumerate}
\item \textbf{Identifier Generalization.}
All identifier placeholders produced during hard template extraction
(e.g., \texttt{table\_name}, \texttt{col\_name}, \texttt{new\_table},
\texttt{new\_column}, \texttt{cte}, and alias tokens such as
\texttt{table\_aliasN} and \texttt{column\_aliasN}) are replaced with a
single token \texttt{variable}. This token represents any table or column
identifier, collapsing different schema roles into a unified symbol. 

\item \textbf{Literal Type Preservation.}
Unlike identifiers, typed literal placeholders introduced in the hard
template (e.g., \texttt{num}, \texttt{string}, \texttt{date},
\texttt{boolean}, \texttt{jsonb}, \texttt{others}) are preserved.
This maintains the semantic role of constants in predicates and clauses
such as \texttt{LIMIT num} or \texttt{variable = string}. 

\item \textbf{Keyword and Structure Retention.}
All SQL keywords, operators, and syntactic structure (e.g.,
\texttt{SELECT}, \texttt{JOIN}, \texttt{GROUP BY}, \texttt{ORDER BY})
remain unchanged so that the generalized template still reflects the
underlying query logic. 
\end{enumerate}

The resulting SQL statement forms a soft template that preserves the structural semantics of the original query.

\section{Frequency Curves for Hard and Soft Templates}
\label{sec:freq}

\begin{figure}[t]
\centering
\begin{subfigure}{\linewidth}
\centering
\includegraphics[width=\linewidth]{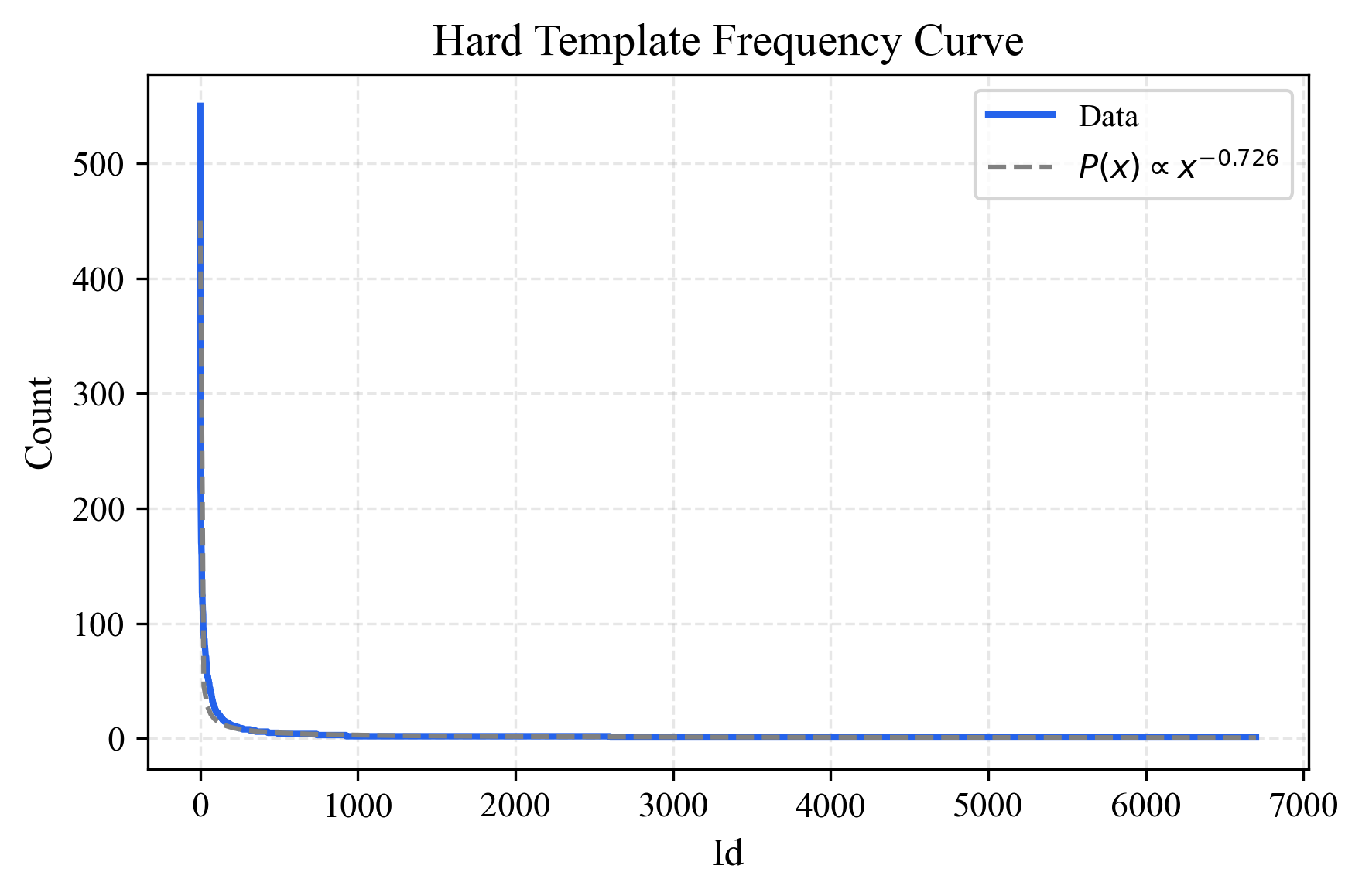}
\caption{Hard templates}
\end{subfigure}

\begin{subfigure}{\linewidth}
\centering
\includegraphics[width=\linewidth]{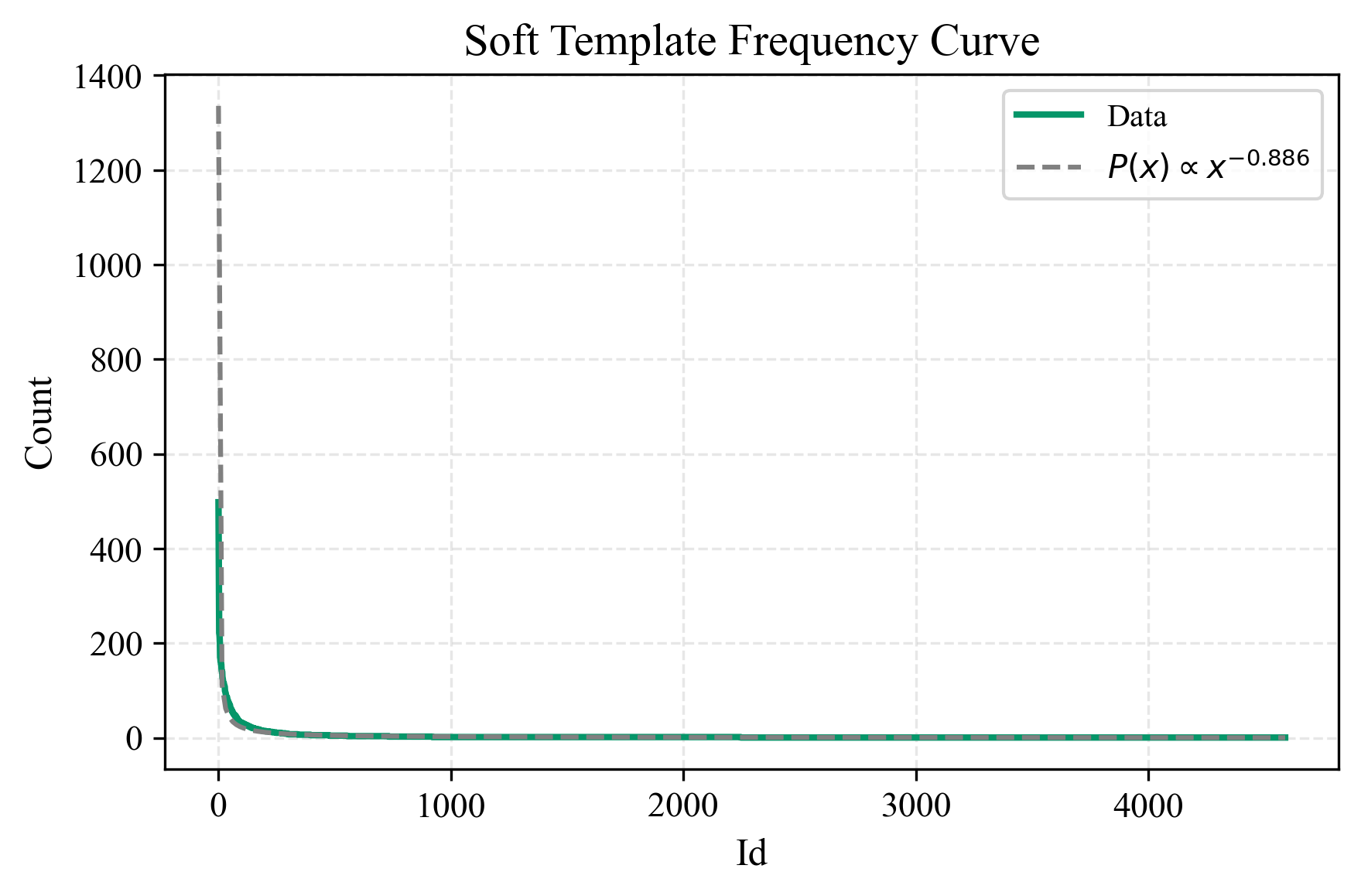}
\caption{Soft templates}
\end{subfigure}

\caption{Frequency curves for hard and soft templates.}
\label{fig:frequency_curve}
\end{figure}


Figure~\ref{fig:frequency_curve} shows the frequency distributions of hard and soft templates, both of which exhibit a Power Law–like pattern. The fitted distributions follow $P(x) \propto x^{-0.726}$ and $P(x) \propto x^{-0.886}$, respectively.

\section{Spearman Correlation Test Results}
Table~\ref{tab:spearman_proxy} shows the Spearman Correlation test results.
\label{appen-Spearman}
\begin{table}[t]
\centering
\small
\caption{Spearman correlation test for the monotonic relationship between each proxy and the table count.} 
\label{tab:spearman_proxy}
\resizebox{\columnwidth}{!}{
\begin{tabular}{lcc}
\toprule
\textbf{Proxy} & \textbf{$\rho$} & \textbf{p-value} \\
\midrule
Num\_tables & -0.3673 & $2.33\times10^{-2}$ \\
Num\_joins & -0.3839 & $1.73\times10^{-2}$ \\
Num\_subqueries & 0.4481 & $4.79\times10^{-3}$ \\
Max\_nesting\_depth & 0.3124 & $5.62\times10^{-2}$ \\
Num\_aggs\_plus\_group\_by & -0.4263 & $7.61\times10^{-3}$ \\
Advanced\_feature\_count & 0.4921 & $1.70\times10^{-3}$ \\
\bottomrule
\end{tabular}}
\end{table}

\section{Individual Plots for Six Proxies}
\begin{figure*}[t]
\centering
\includegraphics[width=0.85\textwidth]{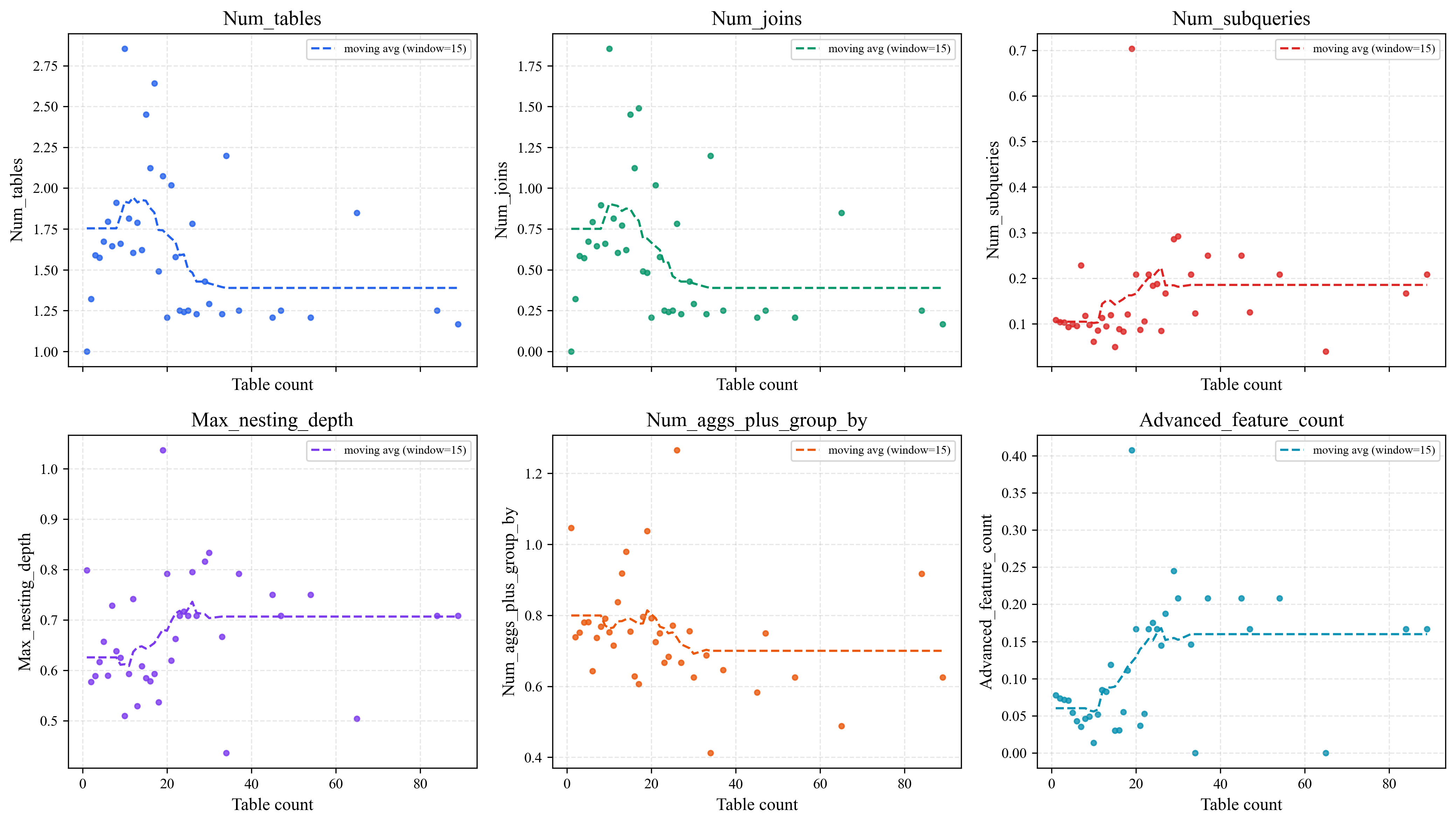}
\caption{Individual Plots for Six Proxies. Each point represents the average value of a proxy metric for a specific table count. The dashed line denotes the moving average of the proxy values with a window size of 15.}
\label{fig:avg_metric} 
\end{figure*}

In figure~\ref{fig:avg_metric}, each point represents the average value of a proxy metric for a specific table count. The dashed line denotes the moving average of the proxy values with a window size of 15.

\section{Average Proxy Values Across template frequency groups.}
\label{appen: avg_metric}

Table~\ref{tab:metric_by_freq} presents the average proxy values across hard and soft templates.
\begin{table}[t]
\centering
\small
\caption{Average proxy values across template frequency groups. The maximum value in each row is highlighted in bold.}
\label{tab:metric_by_freq}

\resizebox{\columnwidth}{!}{
\begin{tabular}{llcccc}
\toprule
\textbf{Metric} & \textbf{Type} & \textbf{High} & \textbf{Middle} & \textbf{Long Tail} & \textbf{Once} \\
\midrule
Num\_tables & Soft & 1.41 & 1.74 & 1.83 & \textbf{2.12} \\
            & Hard & 1.12 & 1.55 & 1.91 & \textbf{2.21} \\
\midrule
Num\_joins & Soft & 0.41 & 0.74 & 0.82 & \textbf{1.10} \\
           & Hard & 0.12 & 0.55 & 0.91 & \textbf{1.20} \\
\midrule
Num\_subqueries & Soft & 0.03 & 0.11 & 0.14 & \textbf{0.23} \\
                & Hard & 0.04 & 0.13 & 0.11 & \textbf{0.15} \\
\midrule
Max\_nesting\_depth & Soft & 0.41 & 0.54 & 0.72 & \textbf{1.06} \\
                    & Hard & 0.43 & 0.54 & 0.64 & \textbf{0.86} \\
\midrule
Num\_aggs\_plus\_group\_by & Soft & 0.49 & 0.64 & 0.94 & \textbf{1.09} \\
                           & Hard & 0.51 & 0.62 & 0.87 & \textbf{0.90} \\
\midrule
Advanced\_feature\_count & Soft & 0.00 & \textbf{0.10} & 0.07 & 0.04 \\
                         & Hard & 0.00 & \textbf{0.10} & 0.07 & 0.02 \\
\bottomrule
\end{tabular}
}
\end{table}

\section{Template Frequency Distribution.}
\label{appen-dist}
Table~\ref{tab:frequency} shows the template frequency distribution. 
\begin{table}[t]
\centering
\small
\caption{Template frequency distribution. Percentages indicate the proportion of templates in each frequency category.}
\label{tab:frequency}
\resizebox{\columnwidth}{!}{
\begin{tabular}{lcccc}
\toprule
\textbf{Template Type} & \textbf{High ($\geq$100)} & \textbf{Middle (10--99)} & \textbf{Long Tail (2--9)} & \textbf{Once (1)}\\
\midrule
Hard & 19 (0.3\%) & 223 (3.3\%) & 2358 (35.2\%) & 4099 (61.2\%) \\
Soft & 28 (0.6\%) & 260 (5.7\%) & 1958 (42.7\%) & 2341 (51.0\%) \\
\bottomrule
\end{tabular}
}
\end{table}

\end{document}